\documentclass[runningheads]{llncs}

\usepackage{times}
\usepackage{latexsym}
\usepackage{bm}
\usepackage{amsmath}
\usepackage{amsfonts}
\usepackage{booktabs}
\usepackage{array}
\usepackage{multirow}
\usepackage{graphicx}
\usepackage{color}
\usepackage{subfigure}
\usepackage{url}
\usepackage{CJKutf8}

\begin{document}

%
%
\title{IPRE: a Dataset for Inter-Personal Relationship Extraction}
\titlerunning{A Dataset for IPRE}
\author{Haitao Wang, Zhengqiu He, Jin Ma, Wenliang Chen, Min Zhang}
\authorrunning{H. Wang et al.}

\institute{School of Computer Science and Technology, Soochow University, China\\
	\email{\{htwang2019,zqhe,jma2018\}@stu.suda.edu.cn,\\
		\{wlchen, minzhang\}@suda.edu.cn}}

\begin{CJK}{UTF8}{gbsn}
\maketitle              
\begin{abstract}

Inter-personal relationship is the basis of human society. In order to automatically identify the relations between persons from texts, we need annotated data for training systems. However, there is a lack of a massive amount of such data so far. To address this situation, we introduce IPRE, a new dataset for inter-personal relationship extraction which aims to facilitate information extraction and knowledge graph construction research. In total, IPRE has over 41,000 labeled sentences for 34 types of relations, including about 9,000 sentences annotated by workers. Our data is the first dataset for inter-personal relationship extraction. Additionally, we define three evaluation tasks based on IPRE and provide the baseline systems for further comparison in future work.

\end{abstract}
\section{Introduction}

Inter-personal relationship, which is the basis of society as a whole, is a strong, deep, or close connection among persons \cite{white2016evolution}. The types of inter-personal relationship include kinship relations, friendship, work, clubs, neighborhoods, and so on. From billions of web pages, we can explore relations of persons to form as knowledge bases (KBs), such as EntityCube (also named as Renlifang) \cite{zhu2009statsnowball} which is a knowledge base containing attributes of people, locations, and organizations, and allows us to discover a relationship path between two persons. 
However, inter-personal relationships in the KBs could be incorrect, which caused by supervised relation extractor because of lack of a massive amount of training data.
 To solve this problem, our task is to build a dataset for training a system to improve the accuracy of inter-personal relationships. 

An obvious solution is to label the data by human annotators. However, hiring the annotators is costly and non-scalable, in terms of both time and money.
To overcome this challenge, Mintz et al. \cite{mintz2009distant} put forward the idea of distant supervision (DS), which can automatically generate training instances via aligning knowledge bases and texts. The key idea of distant supervision is that given an entity pair $<\!\!e_h, e_t\!\!>$ and its corresponding relation $r_B$ from KBs such as Freebase \cite{bollacker2008freebase}, we simply label all sentences containing the two entities by relation $r_B$ \cite{hoffmann2011knowledge,riedel2013relation,surdeanu2012multi}. Several benchmark datasets for distant supervision relation extraction have been developed \cite{mintz2009distant,riedel2010modeling} and widely used by many researchers \cite{he2019syntax,ji2017distant,Lin2016Neural,zeng2015distant}. Through distant supervision, we can easily generate a large scale of annotated data for our task without labor costs. However, the data generated by this way inevitably has wrong labeling problem.
While training noise in distant supervision is expected, noise in the test data is troublesome as it may lead to incorrect evaluation. Ideally, we can easily build a large scale of training data by taking the advantage of distant supervision, and hire annotators to label the testing data to overcome the problem of incorrect evaluations.  


This paper introduces the  Inter-Personal Relationship Extraction (IPRE) dataset in Chinese, in which each entity is a person and each relation is an inter-personal relation, 
e.g., ``姚明" (Yao Ming) and ``叶莉" (Ye Li), and their inter-personal relationship is \emph{wife} (妻子), since Ye Li is the wife of Yao Ming.
The IPRE dataset includes a set of sentence bags, and each bag is corresponding to two persons and their relation, of which the sentences must contain the two persons. In total, the IPRE dataset has over 41,000 sentences grouped into 4,214 bags related to 5,561 persons and 34 relation types. In the data, there are 1,266 manually-annotated bags used as development and test sets, and 2,948 DS-generated bags used as a training set. 



We first define a set of inter-personal relations used in IPRE, which includes 34 types by considering the balance of occurrences in the data. We further present the data collection procedure, a summary of the data structure, as well as a series of analyses of the data statistic. We select some sentences from the collected data into the development and test sets, and annotate them manually.  To show the potential and usefulness of the IPRE dataset, benchmark baselines of distantly supervised relation extraction have been conducted and reported. 
The dataset is available at \url{https://github.com/SUDA-HLT/IPRE}.

Our main contributions are:
\begin{itemize}
	\item To our best knowledge, IPRE is the first dataset for Inter-Personal Relation Extraction.  IPRE can be used for building the systems for identifying the relations among persons and then contribute to construct knowledge base such as EntityCube. 
	\item IPRE can serve as a benchmark of relation extraction. We define three different tasks: Bag2Bag, Bag2Sent, and Sent2Sent (described in Section \ref{sec:bench}) for evaluation. 
	We also provide the baseline systems for the three tasks that can be used for the comparison in future work.
\end{itemize}

\section{Related Work}
In this section, we make a brief review of distant supervision data and human annotated data for relation extraction.

\textbf{Riedel2010} \cite{riedel2010modeling} has been widely used by many researchers \cite{he2019syntax,ji2017distant,Lin2016Neural,zeng2015distant,zhang2018attention}. The dataset uses Freebase as a distant supervision knowledge base and New York Times (NYT) corpus as text resource. Sentences in NYT of the years 2005-2006 are used as training set while sentences of 2007 are used as testing set. There are 52 actual relations and a special relation ``NA" which indicates there is no relation between given entity pair.
The sentences of NA are from the entity pairs that exist in the same sentence of the actual relations but do not appear in the Freebase. 

\textbf{GIDS} \cite{jat2018improving} is a newly developed dataset. To alleviate noise in distant supervision setting, it makes sure that labeled relation is correct and for each instance bag in GIDS, there is at least one sentence in that bag which expresses the relation assigned to that bag. GIDS is constructed from the human-judged Google Relation Extraction corpus, which consists of 5 binary relations: ``perGraduatedFromInstitution", ``perHasDegree", ``perPlaceOfBirth", ``perPlaceOfDeath", and ``NA".


\textbf{ACE05 and CONLL04} \cite{rothYih2004} are the widely used human-annotated datasets for relation extraction. The ACE05 data defines seven coarse-grained entity types and six coarse-grained relation categories, while the CONLL04 data defines four entity types and five relation categories. The test sets of them are about 1,500 sentences, much smaller than ours.

\section{IPRE Dataset}
In this section, we describe how to construct the IPRE dataset. The premise of distant supervision method is that there is a well-organized knowledge base (e.g., Freebase) to provide entity-relation triples. But for inter-personal relations, there is no such knowledge base publicly available so far. Thus in our approach, we should first extract persons-relation triples by ourselves. Then, we generate sentence bags by aligning the triples to the texts in a large scale of text corpus. Finally, we choose a certain percentage of the sentence bags for human annotation.

\subsection{Data Alignment via Distant Supervision}

%
%

\subsubsection{Candidates of Person Entities}
There are several sites containing a large amount of wiki-style pages (e.g., Wikipedia and Chinese Baidu Baike), which can provide enough persons-relation triples. 
We crawl Chinese webpages from the wiki-style sites as our resource to extract persons-relation.

The webpages includes many types of entities besides persons. We need to list the candidates of person entities before extraction. The webpages of an entity usually contains some tags, which are used to describe the category to which the entity belongs. For example, entity ``姚明" (Yao Ming) is tagged as ``运动员 (athlete), 话题人物 (topic character), 篮球运动员 (basketball player), 篮球 (basketball), 体育人物 (sportsman)". From the tags, we can easily figure out that Yao Ming is a person.
We count the number of occurrences of the tags for all entities and manually select the tags with a frequency of more than 100 as the tag set of the category of person. The tag set of person has more than 119 entries, from which top 10 are listed in Table \ref{entity_types}.
 If one of the tags of an entity is included in the tag set, we put the entity into a set of person entities.

Since the webpages might contain some errors involved by the nonprofessional editors, we further clean up the set of person entities. There are three steps: 1) symbol cleanup, all chars should be a Chinese character; 2) multi-source information verification, we check the information in infobox, and the surname should be defined in ``Hundred Family Surnames" \footnote{\url{https://en.wikipedia.org/wiki/Hundred_Family_Surnames}}; 3) length restriction. We limit the length of person entity between 2 and 6 according to relevant regulations about Chinese person name. Finally we obtain a set of 942,344 person entities.





\subsubsection{Inter-Personal Relations}
The types of inter-personal relationship include kinship relations, friendship, marriage, work, clubs, neighborhoods, and so on. 
We first define a set of inter-personal relations (IPR), having 34 types.

\begin{minipage}[bt]{\textwidth}
	\begin{minipage}[t]{0.35\textwidth}
		\centering
		\footnotesize
		\makeatletter\def\@captype{table}\makeatother\caption{Top 10 tags for persons.}
		\label{entity_types}
		\begin{tabular}{|c|l|}
			\hline
			\#1 & 人物 (person) \\\hline
			\#2 & 行业人物 (industry figure) \\\hline
			\#3 & 政治人物 (politician)\\\hline
			\#4 & 教师 (teacher)\\\hline
			\#5 & 学者 (scholar)\\\hline
			\#6 & 体育人物 (sportsman) \\\hline
			\#7 & 官员 (official)\\\hline
			\#8 & 娱乐人物 (entertainment figure)\\\hline
			\#9 & 科研人员 (researcher)\\\hline
			\#10 & 教授 (professor) \\
			\hline
		\end{tabular}
	\end{minipage}
	\begin{minipage}[t]{0.65\textwidth}
		\centering
		\footnotesize
		\makeatletter\def\@captype{table}\makeatother\caption{Relation types in IPR.}
		\label{relation_types}
		\begin{tabular}{|l|l|c|c|}
			\hline
			\multicolumn{1}{|c|}{ID} &  \multicolumn{1}{c|}{Relation Type} & \# Sentence & \# Triple \\
			\hline
			\#1 & 现夫(husband)  & 8896 & 571  \\
			\hline
			\#2 &生父(father)	&7607	&986   \\
			\hline
			\#3 &现妻(wife)	&5965	&363 \\
			\hline
			\#4 &老师(teacher)	&3174	&287  \\
			\hline
			\#5 &儿子(son)	&2936	&294  \\
			\hline
			... & ... & ... & ...\\
			\hline
			\#32 &公公(father-in-law)	&25	&7 \\
			\hline
			\#33 &儿媳(daughter-in-law)	&14	&8  \\
			\hline
			\#34  &外公(grandfather)	&12	&5 \\
			\hline
		\end{tabular}
	\end{minipage}
	\vspace{10pt}
\end{minipage}

From the infobox of the pages, we can obtain more than 1,700 expressions of relations between persons. Some expressions have the same meaning and can be mapped to one entry of IPR. 
We manually map the expressions to the entries of IPR. 
Due to the space limitation, we only list some types in Table \ref{relation_types}.
Based on 34 types and their associated expressions, we can extract persons-relation triples from the infobox.



\subsubsection{Text Alignment}

We follow the general procedure of distant supervision used for relation extraction \cite{mintz2009distant}. 
We make a sentence pool from the texts in the webpages. 
Then we use a name entity tagger to detect the person names in the sentences. 
If one sentence contains two person names which appear in one relation triple, the sentence is selected. All the sentences that contain an entity pair are grouped into one bag. Finally, we have over 41,000 sentences and 4,214 bags.

\subsection{Manually Annotating}
After obtaining the data generated by distant supervision, we divide this dataset into training (70\%), development (10\%) and test (20\%) sets, ensuring that there is no overlap between these sets. 
As we mentioned before, we manually annotate all sentences in the development and test sets for evaluation purpose.
Given a sentence and the matched entity pair ($<e_h, e_t>$), there are three steps: 1) the annotators first check whether the sentence is legal; 2) they check if the mentions of $e_h$ and $e_t$ are person names since some Chinese names could be used as common words; 3) they determine the relation of $e_h$ and $e_t$ expressed by the sentence. If one sentence cannot go through the first two steps, it is an illegal sentence. At step 3), the annotators choose one of 34 types if possible, otherwise use NA as the type. 



Four workers annotate 9,776 sentences manually and each sentence is marked at least twice. 
When the annotations are inconsistent, the third annotator will relabel the data. 
The Kappa coefficient of annotation results is about 0.87, which is a satisfactory value.
In these sentences, there are 899 illegal sentences. 
In the remaining legal sentences, 
3,412 sentences express one relation out of 34 types, while the others are marked as NA. At bag-level, the legal sentences are grouped into 1,266 bags, while 941 bags express at least one relation of 34 types. 




\begin{figure*}[tb]
	\centering
	\setlength{\belowcaptionskip}{-0.35cm} 
	\begin{minipage}{0.30\textwidth}
		\centering
		\includegraphics[width=\textwidth]{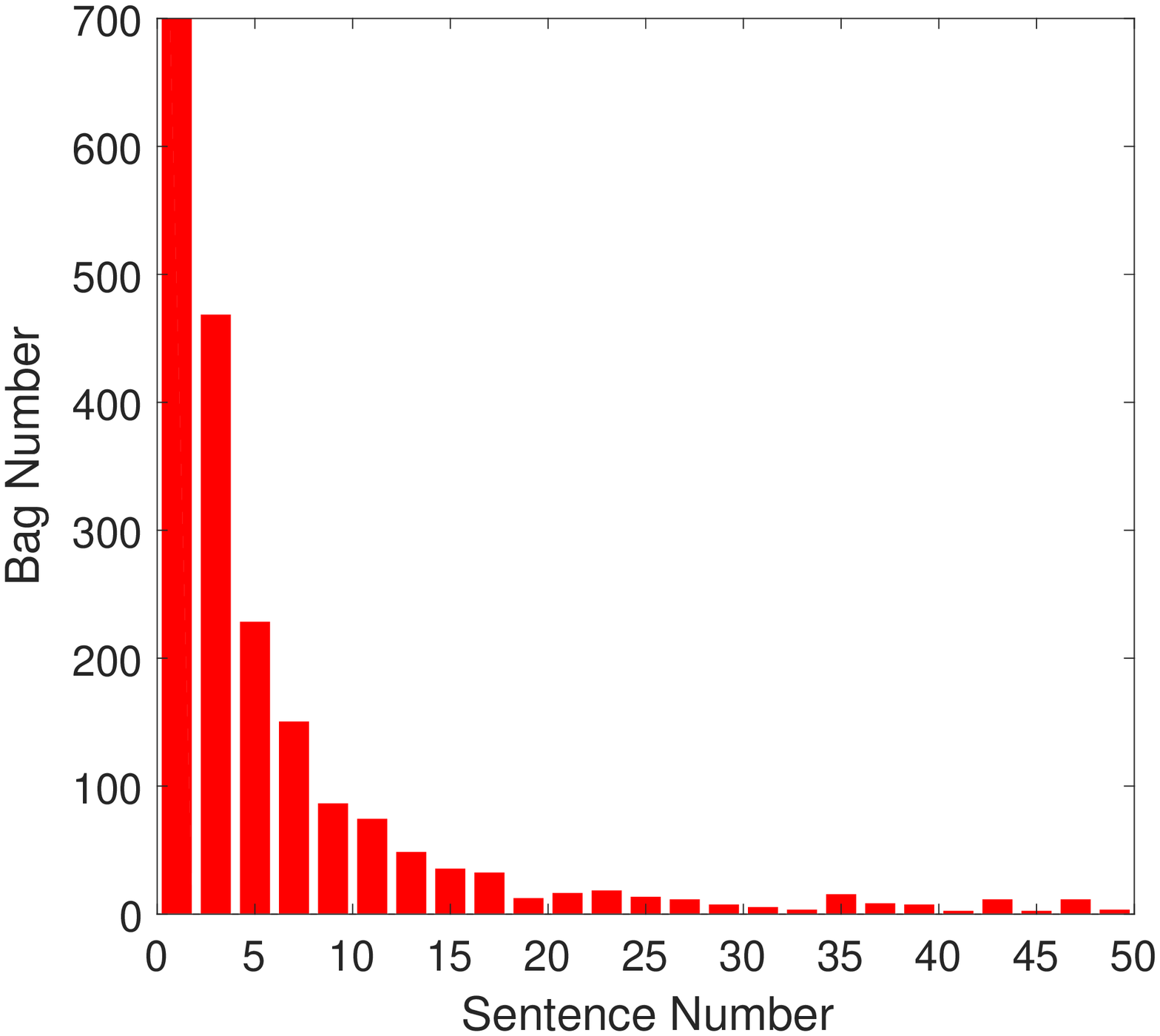}
		\caption{Sentence number of each bag.}
		\label{fig:sentence_in_bag}
	\end{minipage}
	\begin{minipage}{0.30\textwidth}
		\centering
		\includegraphics[width=\textwidth]{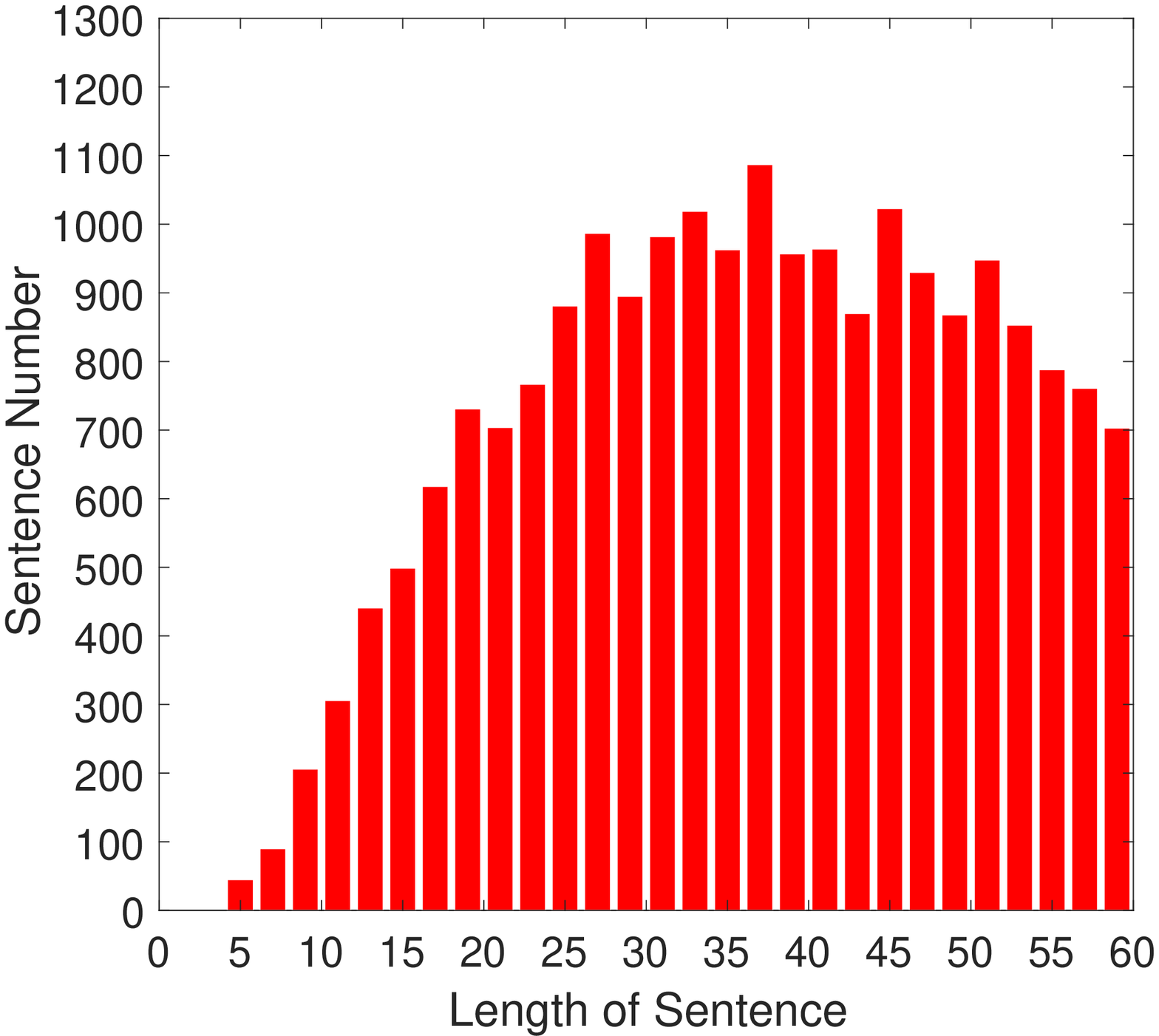}
		\caption{Sentence length distribution.}
		\label{fig:length_distribution_of_sentences}
	\end{minipage}
	\begin{minipage}{0.30\textwidth}
		\centering
		\includegraphics[width=\textwidth]{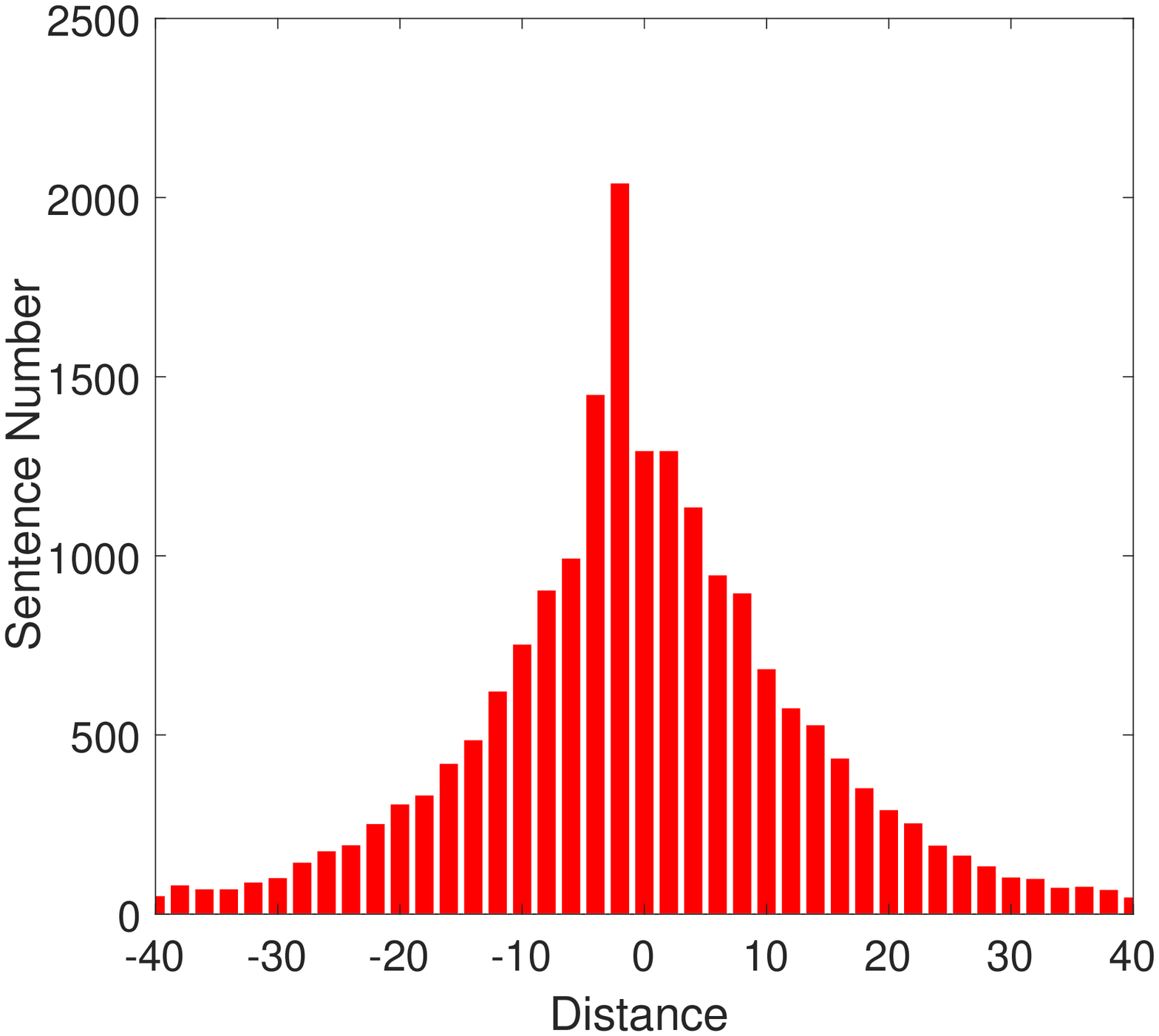}
		\caption{Distance between entity pair.}
		\label{fig:entity_distance}
	\end{minipage}
\end{figure*}

\subsection{Data Statistics}

In order to have a more comprehensive and clear understanding of IPRE, we have statistics on the relevant features of IPRE. Our goal is to construct a dataset with a relatively balanced amount of data among types, so we first count the number of triples for each type and the number of sentences matched. Table \ref{relation_types} shows the statistical results, from which we can see that the majority of the categories have more than 10 triples.

Figure \ref{fig:sentence_in_bag} further gives the distribution histogram of the number of sentences contained in each bag. Although there are some bags with only one sentence, the number of sentences in most bags is between 2 and 15, which is a very reasonable distribution. 

Figure \ref{fig:length_distribution_of_sentences} shows the distribution of sentence length. 
In order to keep quality of sentences as possible, 
we limit the length of the sentence to a maximum of 60 in the process of text alignment section. From the figure we can see that the sentence length is mostly more than 15, and it can form a complete semantic unit to a large extent. 

Meanwhile, many previous studies have shown that the closer the distance between two entities in the same sentence, the more likely it is that the sentence will reflect the semantic relationship of the entity pair \cite{Lin2016Neural,Zeng2014Relation,zeng2015distant}. From Figure \ref{fig:entity_distance}, we can see that in the IPRE dataset, the distance between two entities in most sentences is between 3-20 words,  which is a reasonable distance that can reflect the semantic relationship.




\section{IPRE as a New Benchmark}\label{sec:bench}

\begin{figure} [tb]
	\centering
	\setlength{\belowcaptionskip}{-0.5cm} 
	\begin{minipage}{0.5\textwidth}
		\centering
		\includegraphics[width=\textwidth]{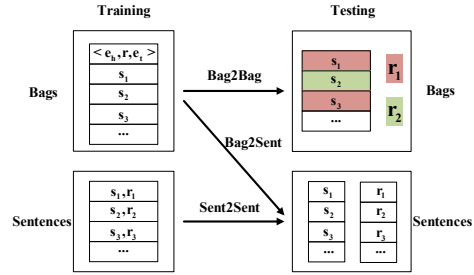}
	\end{minipage}
	\caption{Three types of task based on IPRE.}
	\label{fig:three_task}
\end{figure}

Based on IPRE,
we design relation extraction tasks from three types of learning strategies.
The first one is the most commonly used learning paradigm at bag-level in distant supervision data, namely multi-instance learning (MIL). 
  Here, all sentences corresponding to a relational triple are regarded as a bag,
and the relation prediction is at bag-level during both training and testing.
The second one is to employ bag-level method as we do in the first one.
And since the test set has been labeled manually, it is able to predict the relation at sentence-level when testing.
The third one is to treat it as a general task of  relation extraction, and to train and
predict relation at sentence-level. Figure \ref{fig:three_task} shows the three tasks. 

\subsection{Bag2Bag: Bag-Training and Bag-Testing}
In order to mitigate the impact of the wrong labeling problem in distant supervision data, 
the existing work mainly focuses on MIL \cite{Lin2016Neural,zeng2015distant}. 
However, in the existing corpus constructed by distant supervision,
the training set and test set are processed in the same way.
That is, the only supervised information is the entity-relation triples 
existing in KBs, so it is uncertain whether one sentence in a bag 
is labeled correctly or not.
For this reason, what may happen when verify the performance of the relation extraction
system at the testing phase is that although an entity-relation is correctly predicted,
it  is based on sentences that do not actually belong to the relation type.

In real-world scenarios, there is a need for bag-level training and testing. 
For example, when there is a lot of noise data in the constructed corpus, 
or we only need to determine the relation type between entities, but do not care
about the expression of specific sentences.
Based on IPRE, we can  make more stringent bag-level relation prediction, as illustrated in Figure \ref{fig:three_task}.
Since we provide the labeling results of all sentences in the development set and 
test set, we can not only predict the relation types of bags, but also have stricter requirements for the predicted results that the system should output the sentence or sentences in the bag
that support the relation prediction. This is more reasonable than the black-box-level
bag relation prediction in the existing work.

Here, we give the detail evaluation metrics of Bag2Bag task.
In distant supervision data, the number of samples of NA relation is usually very large.
Therefore, NA relation is generally not considered in the assessment.
Let $t_p$ be the true positive bags, $f_p$ be the false positive bags, $f_n$ be the false negative bags.
Thus, we have:
\begin{equation}
\label{eq:bag2bag}
\begin{split}
P   = \frac{I_{t_p}}{I_{t_p}+I_{f_p}}, 
R   = \frac{I_{t_p}}{I_{t_p}+I_{f_n}}, 
F_1   = \frac{2PR}{P+R}
\end{split}
\end{equation}
where $I$ is the counting function.
It is worth noting that  the number of each bag is equal to the number of non-NA relations assigned to this bag when counting the number of bags.


\subsection{Bag2Sent: Bag-Training and Sentence-Testing}
There is another requirement
scenario that we need to train relation extractor at bag-level because there are many
noises in the given corpus, or we are looking for more stable model training. However, in 
the follow-up model application, we need to determine the relation between entities
in each specific sentence. Therefore, it is necessary to predict the relation at 
sentence-level. 

Based on IPRE, we can employ bag-level training method in model training process
by treating each bag as a semantically enriched sentence, and extract the features of the sentences in the
bag that are closely related to the relation prediction of a given entity pair. At the same
time, since each sentence is labeled manually in the development set and test set, the relation
prediction of a single sentence can be made.

Let $t_p$ be the true positive sentences, $f_p$ be the false positive sentences, $f_n$ be the false negative sentences.  
NA relation is also not considered like Bag2Bag, and the evaluation metrics of this task is formally the same as Equation (\ref{eq:bag2bag}).

\subsection{Sent2Sent: Sentence-Training and Sentence-Testing}
In the research of relation extraction task, more attention is focused on the training and
prediction of single sentence. 
For distant supervision relation extraction,
when we construct training data with fewer errors,
or the proposed model has better anti-noise ability,
or we have the ability to filter the noise in training data, 
such as the recently proposed method of using 
reinforcement learning to select sentences \cite{feng2018rl},
it will be a good practice to  make sentence-level
relation extraction from distant supervision data.

IPRE is a good cornerstone for the research of sentence-level relation extraction task.
We can make full use of the characteristics of the training set containing noise data to 
design more robust and noise-resistant models, and we can also design better strategies to
identify noise data. Meanwhile, the fully annotated development set and test set can achieve
the purpose of measuring the performance of the model.

The Sent2Sent task shares the same strategy with the Bag2Sent and also uses the Equation (\ref{eq:bag2bag}) to evaluate the  prediction results at sentence-level, as shown in Figure \ref{fig:three_task}.



\section{Baselines}
\label{baselines}
In this section, we present several baseline models for 
three defined tasks. We first describe three sentence encoders. Then, based on the sentence encoders we build the systems for each task.


\subsection{Sentence Encoder}
The sentence encoders transform a sentence into its distributed representation. 
First, the words in the input sentence are embedded into dense real-valued feature vectors.
Next, the neural networks are used to construct the distributed representation of the sentence. 

\subsubsection{Input Representation}
Given a sentence $ s = \{w_1, ..., w_n \}$, where $w_i$ is the $i$-th word in the sentence, the input is a matrix composed of $n$ vectors
$\bm{x} = [\bm{x_1}, ..., \bm{x_n}]$, where $\bm{x_i}$ corresponds to $w_i$ and consists of its word embedding and  position embedding. Following previous work \cite{Lin2016Neural,zeng2015distant}, we employ the skip-gram method \cite{mikolov2013b} to 
 the word embedding. 
Position embeddings are first successfully applied to relation extraction task by Zeng et al. \cite{Zeng2014Relation}, which specifies the relative distances of a word with respect to two target entities. 
In this way, $\bm{x_i} \in \mathbb{R}^d $, and $d = d^a + 2 \times d^b$, where $d^a$ and $d^b$ are the dimensions of word embedding and position embedding respectively.

\subsubsection{CNN/PCNN Encoder}
After encoding the input words, a convolution layer is applied to reconstruct the original input $\bm{x}$ by learning sentence features from a small window of words at a time while preserving word order information. We use $K$ convolution filters with the same window size $l$.
Then, we combine the $K$ convolution output vectors via a max-pooling operation to obtain a vector of a fixed length $K$.
Formally, the $j$-th element of the output vector $\bm{c} \in \mathbb{R}^{K}$ as follows:
\begin{equation}
\label{eq:con_maxpool}
\begin{split}
& \bm{c}[j] = \mathop{\mathrm{max}}\limits_{i}(\bm{W}_j\bm{x}_{i:i+l-1}+\bm{b}_j)
\end{split}
\end{equation}
where $\bm{W}_j$ and $\bm{b}_j$ are model parameters. 



Further, Zeng et al. \cite{zeng2015distant} adopts piecewise max pooling (PCNN) in relation extraction, which is a variation of CNN. Suppose the positions of the two entities are $p_1$ and $p_2$ respectively. Then, each convolution output vector is divided into three segments:
\begin{equation}\label{cnn_piecewise}
\begin{split}
& [0:p_1-1],[p_1:p_2],[p_2+1:n-l]
\end{split}
\end{equation}

The max scalars in each segment are preserved to form a 3-element vector, and all vectors produced by the $K$ filters are concatenated into a  vector with length $3K$, which is the output of the pooling layer.

Finally, we apply a non-linear transformation (e.g., tanh) on the output vector to obtain the sentence embedding $\bm{s}$.

\subsubsection{Bi-LSTM Encoder}

LSTM units \cite{hochreiter1997long} can keep the previous state and memorize the extracted features of the current data input.
Following Zhou et al. \cite{zhou2016attention}, we adopt a variant introduced by Graves et al. \cite{graves2013speech}, and each unit is 
computed as:
\begin{equation}\label{bilstm}
\setlength{\abovedisplayskip}{1pt}
\setlength{\belowdisplayskip}{1pt}
\begin{split}
\bm{h_{t}},\bm{c_{t}} = LSTM(\bm{h_{t-1}}, \bm{c_{t-1}})
\end{split}
\end{equation}
In this paper, we use  bidirectional LSTM networks (Bi-LSTM) \cite{graves2005framewise} to encode the sentences. 
Therefore, the forward and backward outputs of the $i$-th word are concatenated:
\begin{equation}
\setlength{\abovedisplayskip}{1pt}
\setlength{\belowdisplayskip}{1pt}
\label{concat}
\begin{split}
& \bm{h}_i = [\overrightarrow{\bm{h}_i}; \overleftarrow{\bm{h}_i}] \\
\end{split}
\end{equation}

Then, an attention layer is exploited to determine which segments in a sentence are most influential. More specifically, for the matrix $\bm{H} = \{\bm{h}_1, \bm{h}_2, ..., \bm{h}_n\}$, we compute its attention vector as:
\begin{equation}
\setlength{\abovedisplayskip}{-1pt}
\label{word-attention}
\begin{split}
& \bm{\alpha} = \mathrm{softmax}(\bm{\omega}^T\mathrm{tanh}(\bm{H})) \\
\end{split}
\end{equation}
where $\bm{\omega}$ is a trained parameter vector and $\bm{\omega}^T$ is a transpose. Next, we can obtain the sentence representation as:
\begin{equation}
\setlength{\abovedisplayskip}{-1pt}
\setlength{\belowdisplayskip}{-1pt}
\label{weight_sum}
\begin{split}
& \bm{s} = \mathrm{tanh}(\bm{H}\bm{\alpha}^T)
\end{split}
\end{equation}

\subsection{Systems for Bag2Bag Task}
\label{sec:bt-bt}

For the Bag2Bag task, we employ two MIL methods to perform bag-level
relation prediction, i.e., expressed at least once (ONE)\cite{zeng2015distant},
and  attention over instances (ATT) \cite{Lin2016Neural}.
Suppose that there are $T$ bags $\{B_1, B_2, ..., B_T\}$ and that the $i$-th bag contains $m_i$ instances $B_i = \{s_i^1, s_i^2, ..., s_i^{m_i}\}$. The objective of MIL is to predict the labels of the unseen bags.
\subsubsection{ONE}
Given an input instance $s_i^j$, the network with the parameter $\theta$ outputs a vector $\bm{o}$. To obtain the conditional probability $p(r|s,\theta)$, a softmax operation over all relation types is applied:
\begin{equation}
\setlength{\abovedisplayskip}{1pt}
\setlength{\belowdisplayskip}{1pt}
\label{soft_prob}
\begin{split}
& p(r|s_i^j,\theta) = \frac{\mathrm{exp}(\bm{o}_r)}{\sum_{k=1}^{N_r}{\mathrm{exp}(\bm{o}_k)}}
\end{split}
\end{equation}
where $N_r$ is number of relation types.

Then, the objective function using cross-entropy at  bag-level is defined as follows:
\begin{equation}
\label{one_loss}
\begin{split}
& J(\theta) = \sum_{i=1}^{T}\mathrm{log} \,p(r_i|s_i^j,\theta) \\
& j^* = \mathop{\mathrm{argmax}}\limits_{j} p(r_i|s_i^j,\theta) \\
\end{split}
\end{equation}

Here, when predicting, a bag is positively labeled if and only if the output of the model on at least one of its instances is assigned a positive label.

\subsubsection{ATT}
To exploit the information of all sentences, the ATT model represents the bag $B_i$ with a real-valued vector $\mathbf{emb}_{B_i}$ when predicting relation $r$. It is straightforward that the representation of the bag $B_i$ depends on the representations of all sentences in it. And a selective attention is defined to de-emphasize the noisy sentences.
Thus, we can obtain the representation vector $\mathbf{emb}_{B_i}$ of bag $B_i$ as:
\begin{equation}
    \mathbf{emb}_{B_i} = \sum_j{\frac{\mathrm{exp}(e_j)}{\sum_{k}{\mathrm{exp}(e_k)}}\bm{s}^j_i}
\end{equation}
where $e_j$  scores how well the input sentence $s^j_i$ and the predict relation $r$ matches. Following Lin et al. \cite{Lin2016Neural}, we select the bi-linear form:
\begin{equation}
\label{att_score}
\begin{split}
& e_j = \bm{s}^j_i\bm{A}\bm{r}
\end{split}
\end{equation}
where $\bm{A}$ is a weighted diagonal matrix, and $\bm{r}$ is the 
representation of relation $r$.


After that, the conditional probability $p(r|B_i,\theta)$ through a softmax layer as follows:
\begin{equation}
\label{con_prob}
\begin{split}
\bm{o} & = \bm{M}\mathbf{emb}_{B_i} + \bm{b}\\
 p(r|B_i,\theta) &= \frac{\mathrm{exp}(\bm{o}_r)}{\sum_{k=1}^{N_r}{\mathrm{exp}(\bm{o}_k)}}
\end{split}
\end{equation}
where $\bm{M}$ is the representation matrix of relations and $\bm{b}$ is a bias vector.

Finally, the objective function using cross-entropy at the bag level is defined as:
\begin{equation}
\label{att_loss}
\begin{split}
& J(\theta) = \sum_{i=1}^{T}\mathrm{log} p(r_i|B_i,\theta) \\
\end{split}
\end{equation}


\subsection{Systems for Bag2Sent Task}
For relation extraction task of bag-training and sentence-testing, 
we simply treat each sentence in the test set as a bag, and then as a Bag2Bag task.
Therefore, we still train the model in the same way as Section \ref{sec:bt-bt}. When testing,
we separate each sentence into a bag and make relation prediction.

\subsection{Systems for Sent2Sent Task}
For Sent2Sent task, 
after we get the  representation of sentence $s$, we apply a  MLP to output the confidence vector $\bm{o}$. Then the conditional probability of $i$-th relation is:
\begin{equation}
\label{sen_prob}
\begin{split}
& p(r_i|s,\theta) = \frac{\mathrm{exp}(\bm{o}_i)}{\sum_{k=1}^{N_r}{\mathrm{exp}(\bm{o}_k)}}
\end{split}
\end{equation}

\section{Experiments}

In this section, we present the experimental results and detailed analyses. To evaluate the effect of Chinese word segmentation, we provide two different inputs: word-level and char-level. For word-level, we use Jieba\footnote{\url{https://github.com/fxsjy/jieba}} to perform word segmentation. The evaluation metrics are described in Section \ref{sec:bench}. We report F1 values in the following experiments.

\begin{table}[t]
	\centering
	\small
	\caption{Results of three tasks}
	\label{tb:result_of_three_tasks}
	\begin{tabular}{|c|c|c|c|c|c|c|c|}
		\hline
		\multicolumn{2}{|c|}{Types} & \multicolumn{2}{c|}{Bag2Bag} & \multicolumn{2}{c|}{Bag2Sent} & \multicolumn{2}{c|}{Sent2Sent} \\ \hline
		\multicolumn{2}{|c|}{Method} & \ \ word\ \  &\ \  char\ \  & \ \ word \ \ & \ \ char\ \  &\ \  word\ \  &\ \  char\ \  \\ \hline
		\multirow{2}{*}{CNN} &\  ONE\  & 0.352 & 0.328 & 0.205 & 0.200 & \multirow{2}{*}{0.241} & \multirow{2}{*}{0.160} \\ \cline{2-6}
		&\  ATT\  & 0.359 & 0.328 & 0.220 & 0.214 &  &  \\ \hline
		\multirow{2}{*}{PCNN} &\  ONE\  & 0.291 & 0.287 & 0.149 & 0.162 & \multirow{2}{*}{0.215} & \multirow{2}{*}{0.156} \\ \cline{2-6}
		&\  ATT\  & 0.305 & 0.281 & 0.142 & 0.147 &  &  \\ \hline
		\multirow{2}{*}{\ Bi-LSTM\ } &\  ONE\  & 0.280 & 0.301 & 0.168 & 0.174 & \multirow{2}{*}{0.237} & \multirow{2}{*}{0.186} \\ \cline{2-6}
		&\  ATT \ & 0.296 & 0.334 & 0.157 & 0.171 &  &  \\ \hline
	\end{tabular}
\end{table}

\subsection{Hyperparameter Settings}
In the experiments, we tune the hyperparameters of all the methods on the training dataset and development dataset.
The dimension of word embedding $d^a$ is set to 300, the dimension of position embedding $d^b$ is set to 5, the number of filters $K$ and the dimension of hidden state are set to 300, and the window size $l$ of filters is 3. The batch size is fixed to 50, the dropout \cite{srivastava2014dropout} probability is set to 0.5. When training, we apply Adam \cite{kingma2014adam} to optimize parameters, and the learning rate is set to 0.001.

\subsection{Experimental Results}
Table \ref{tb:result_of_three_tasks} shows the F1 values of different models mentioned in Section \ref{baselines}. To demonstrate the performance of different sentence representation models in Bag2Bag task, we use F1 value defined in Equation (\ref{eq:bag2bag}) for the evaluation criterion. 

From the table, we have the following observation: 
(1) Among the models of sentence semantic representation, CNN is still the most outstanding.
In the task of inter-personal relation extraction, especially on this dataset, local features are usually sufficient to express the relation between persons, and CNN can capture these local features effectively.
(2) For systems based on both PCNN and Bi-LSMT with char-level in the task of Bag2Sent, they perform better than the ones with word-level while the systems of CNN have a different trend. This indicates that for bag-level training, the word information has an uncertain effect.
(3) Compared with task of Bag2Bag and Bag2Sent, the char-level models have lower scores than the word-level models for all three methods in the task of Sent2Sent. This indicates that word embedding has better robustness in noisy data, which may be caused by the fact that words are the most basic semantic unit in Chinese.

\section{Conclusion}

In this paper we introduce the IPRE dataset, a data of over 41,000 labeled sentences (including about 9,000 manually-annotated sentences) for inter-personal relationship. Our data is the first data for extracting relations between persons and much larger than other human-annotated data (such as ACE05 and CONLL04), in terms of both the numbers of sentences and relation types in test sets. Additionally, IPRE can serve as a benchmark of relation extraction. We define three evaluation tasks based on IPRE and provide the baseline systems for further comparison.
\section{Acknowledgements}
The research work is supported by the National Natural Science Foundation of China (Grant No. 61572338, 61876115) and Provincial Key Laboratory for Computer Information Processing Technology, Soochow University. Corresponding author is Wenliang Chen. We would also thank the anonymous reviewers for their detailed comments, which have helped us to improve the quality of this work.

\bibliographystyle{splncs04}
\bibliography{references}

\begin{thebibliography}{10}
\providecommand{\url}[1]{\texttt{#1}}
\providecommand{\urlprefix}{URL }
\providecommand{\doi}[1]{https://doi.org/#1}

\bibitem{bollacker2008freebase}
Bollacker, K., Evans, C., Paritosh, P., Sturge, T., Taylor, J.: Freebase: a
  collaboratively created graph database for structuring human knowledge. In:
  Proceedings of SIGMOD. pp. 1247--1250 (2008)

\bibitem{feng2018rl}
Feng, J., Huang, M., Zhao, L., Yang, Y., Zhu, X.: Reinforcement learning for
  relation classification from noisy data. In: Proceedings of AAAI. pp.
  5779--5786 (2018)

\bibitem{graves2013speech}
Graves, A., Mohamed, A.r., Hinton, G.: Speech recognition with deep recurrent
  neural networks. In: Proceedings of ICASSP. pp. 6645--6649. IEEE (2013)

\bibitem{graves2005framewise}
Graves, A., Schmidhuber, J.: Framewise phoneme classification with
  bidirectional lstm and other neural network architectures. Neural Networks
  \textbf{18}(5-6),  602--610 (2005)

\bibitem{he2019syntax}
He, Z., Chen, W., Li, Z., Zhang, W., Shao, H., Zhang, M.: Syntax-aware entity
  representations for neural relation extraction. Artificial Intelligence
  (2019)

\bibitem{hochreiter1997long}
Hochreiter, S., Schmidhuber, J.: Long short-term memory. Neural computation
  \textbf{9}(8),  1735--1780 (1997)

\bibitem{hoffmann2011knowledge}
Hoffmann, R., Zhang, C., Ling, X., Zettlemoyer, L., Weld, D.S.: Knowledge-based
  weak supervision for information extraction of overlapping relations. In:
  Proceedings of ACL. pp. 541--550 (2011)

\bibitem{jat2018improving}
Jat, S., Khandelwal, S., Talukdar, P.: Improving distantly supervised relation
  extraction using word and entity based attention. arXiv preprint
  arXiv:1804.06987  (2018)

\bibitem{ji2017distant}
Ji, G., Liu, K., He, S., Zhao, J.: Distant supervision for relation extraction
  with sentence-level attention and entity descriptions. In: Proceedings of
  AAAI. pp. 3060--3066 (2017)

\bibitem{kingma2014adam}
Kingma, D.P., Ba, J.L.: Adam: Amethod for stochastic optimization. In:
  Proceedings of ICLR (2014)

\bibitem{Lin2016Neural}
Lin, Y., Shen, S., Liu, Z., Luan, H., Sun, M.: Neural relation extraction with
  selective attention over instances. In: Proceedings of ACL. pp. 2124--2133
  (2016)

\bibitem{mikolov2013b}
Mikolov, T., Sutskever, I., Chen, K., Corrado, G.S., Dean, J.: Distributed
  representations of words and phrases and their compositionality. In:
  Proceedings of NIPS. pp. 3111--3119 (2013b)

\bibitem{mintz2009distant}
Mintz, M., Bills, S., Snow, R., Jurafsky, D.: Distant supervision for relation
  extraction without labeled data. In: Proceedings of ACL. pp. 1003--1011
  (2009)

\bibitem{riedel2010modeling}
Riedel, S., Yao, L., McCallum, A.: Modeling relations and their mentions
  without labeled text. Machine learning and knowledge discovery in databases
  pp. 148--163 (2010)

\bibitem{riedel2013relation}
Riedel, S., Yao, L., McCallum, A., Marlin, B.M.: Relation extraction with
  matrix factorization and universal schemas. In: Proceedings of HLT-NAACL. pp.
  74--84 (2013)

\bibitem{rothYih2004}
Roth, D., Yih, W.T.: A linear programming formulation for global inference in
  natural language tasks. In: Proceedings of CoNLL. pp.~1--8 (2004)

\bibitem{srivastava2014dropout}
Srivastava, N., Hinton, G.E., Krizhevsky, A., Sutskever, I., Salakhutdinov, R.:
  Dropout: a simple way to prevent neural networks from overfitting. Journal of
  Machine Learning Research  \textbf{15}(1),  1929--1958 (2014)

\bibitem{surdeanu2012multi}
Surdeanu, M., Tibshirani, J., Nallapati, R., Manning, C.D.: Multi-instance
  multi-label learning for relation extraction. In: Proceedings of EMNLP. pp.
  455--465 (2012)

\bibitem{white2016evolution}
White, L.A.: The evolution of culture: the development of civilization to the
  fall of Rome. Routledge (2016)

\bibitem{Zeng2014Relation}
Zeng, D., Liu, K., Lai, S., Zhou, G., Zhao, J.: Relation classification via
  convolutional deep neural network. In: Proceedings of COLING. pp. 2335--2344
  (2014)

\bibitem{zeng2015distant}
Zeng, D., Liu, K., Chen, Y., Zhao, J.: Distant supervision for relation
  extraction via piecewise convolutional neural networks. In: Proceedings of
  EMNLP. pp. 1753--1762 (2015)

\bibitem{zhang2018attention}
Zhang, N., Deng, S., Sun, Z., Chen, X., Zhang, W., Chen, H.: Attention-based
  capsule network with dynamic routing for relation extraction. In: Proceedings
  of EMNLP. pp. 986--992 (2018)

\bibitem{zhou2016attention}
Zhou, P., Shi, W., Tian, J., Qi, Z., Li, B., Hao, H., Xu, B.: Attention-based
  bidirectional long short-term memory networks for relation classification.
  In: Proceedings of ACL. vol.~2, pp. 207--212 (2016)

\bibitem{zhu2009statsnowball}
Zhu, J., Nie, Z., Liu, X., Zhang, B., Wen, J.R.: Statsnowball: a statistical
  approach to extracting entity relationships. In: Proceedings of WWW. pp.
  101--110. ACM (2009)

\end{thebibliography}

\end{CJK}
\end{document}